\documentclass[runningheads]{llncs}
\usepackage[T1]{fontenc}
\usepackage{graphicx}
\usepackage[nolist]{acronym}
\usepackage[ruled,vlined]{algorithm2e}
\usepackage{amsmath} 
\usepackage{amssymb}
\usepackage{soul,xcolor}
\usepackage{xcolor}
\begin{acronym}
    \acro{CFD}{Computational Fluid Dynamic}
    \acro{GMRF}{Gaussian Markov Random Field}
    \acro{GSL}{Gas Source Localization}
    \acro{IPP}{Informative Path Planning}
    \acro{MAE}{Mean Absolute Error}
    \acro{MCS}{Motion Capture System}
    \acro{MOX}{Metal OXide}
    \acro{PDE}{Partial Differential Equation}
    \acro{PID}{Photo-Ionization detector}
    \acro{PCB}{Printed Circuit Board}
    \acro{RMSE}{Root Mean Square Error}
    \acro{ROS}{Robot Operating System}
    \acro{STE}{Source Term Estimation}
    \acro{ADE}{Advection-Diffusion Equation}
    \acro{DDPM}{Data-Driven Plume Model}
    \acro{FEM}{Finite Element Method}
    \acro{FVM}{Finite Volume Method}
    \acro{FDM}{Finite Difference Method}
    \acro{GKM}{Gaussian Kernel Method}
    \acro{PDF}{Probability Density Function}
    \acro{DBSCAN}{Density-Based Spatial Clustering of Applications with Noise}
    \acro{NS}{Navier-Stokes}
    \acro{MAP}{Maximum a Posteriori}
    \acro{PoE}{Product of Experts}
    \acro{SLAM}{Simultaneous Localization and Mapping}
    \acro{MRS}{Multi-robot System}
    \acro{CNN}{Convolutional Neural Network}
    \acro{EDF}{Empirical Distribution Function}
\end{acronym}
\begin{document}
\title{Probabilistic Multi-Robot Gas Source Localization with Uncalibrated Sensors: A Distributed Estimation Approach}
\titlerunning{Distributed Multi-Robot Gas Source Localization with Uncalibrated Sensors}
%
\author{Wanting Jin \and
Marc Zoel Arias Mitjà \and
Alcherio Martinoli}
\authorrunning{W. Jin et al.}
%
\institute{ Distributed Intelligent Systems and Algorithms Laboratory
\newline School of Architecture, Civil and Environmental Engineering 
\newline \'{E}cole Polytechnique F\'{e}d\'{e}rale de Lausanne (EPFL), Switzerland
\email{wanting.jin@epfl.ch, mariasmitja@gmail.com, alcherio.martinoli@epfl.ch }}

\maketitle              
\begin{abstract}
Estimating environmental states with multi-robot systems becomes particularly challenging when robots are equipped with uncalibrated and therefore heterogeneous sensors, whose nonlinear and inconsistent responses prevent reliable information fusion. In this paper, we propose a distributed probabilistic framework for source localization tasks that enables calibration-free estimation in the presence of sensor heterogeneity. The key idea is that each robot independently estimates a local belief using a rank-based feature that captures the relative evolution of observations and is invariant to sensor scaling and nonlinearities. These local beliefs are then fused through a product of experts formulation to obtain a consistent global estimate across the team. To further improve the efficiency of team coordination, we introduce an informative region allocation and path planning strategy that reduces redundant exploration while balancing exploration and exploitation. We validate the proposed framework using high-fidelity simulations with realistic gas sensor models. Results demonstrate that our method significantly outperforms a benchmark method based on standard measurement aggregation, achieving reliable source localization accuracy despite strong sensor heterogeneity. More broadly, this work demonstrates how calibration-free sensing representations can be effectively extended to distributed robotic systems, paving the way for their application to other estimation tasks involving heterogeneous sensors.

\end{abstract}
\section{Introduction}
Estimating environmental states from partial observation collected by a \ac{MRS} is an active research area in robotics, with applications in \ac{SLAM}~\cite{tian2022kimera}, moving target tracking~\cite{jung2006cooperative}, and source localization, including sound~\cite{rascon2017localization}, radiation\cite{werner2024autonomous}, and gas sources\cite{Miao_Wang_2024}. These tasks are often time-sensitive, particularly in emergency scenarios, where rapid and accurate estimation is critical. A \ac{MRS} can significantly speed up the estimation process by concurrently gathering data at multiple locations. 

However, practical deployments of a MRS introduce two key challenges. First, robots are often equipped with heterogeneous sensors whose responses can differ significantly due to manufacturing variability and nonlinear characteristics. This is particularly evident in gas sensing, where in-situ sensors exhibit diverse sensitivities and nonlinear responses to concentration levels. In addition, their sensitivity is influenced by temperature, humidity, and sensor age, which makes them difficult to remain calibrated for a long time. Second, effective coordination is required to ensure that robots collect complementary information while avoiding redundant exploration. In this paper, we tackle the \textbf{information sharing} and \textbf{coordinated active perception} for a \ac{MRS} endowed with heterogeneous gas sensors and carrying out a \ac{GSL} task.

Probabilistic approaches have been widely used for \ac{GSL} tasks~\cite{park2023information}, where the belief over candidate source locations is updated by comparing measured gas concentrations with predictions from gas plume models. In multi-robot settings, existing methods \cite{bourne_decentralized_2020,Ercolani_multi_robot_2024,park_cooperative_2020,rahbar_distributed_2020,wiedemann_model-based_2019} typically aggregate measurements directly from all robots to estimate a shared belief. However, these approaches often assume homogeneous sensors or rely on simple noise models, which limit their applicability in realistic scenarios with heterogeneous sensing responses. In \cite{nanavati_distributed_2024}, a distributed estimation has been explored where robots maintain individual beliefs and gradually converge to a consensus. Nevertheless, it still assumes comparable sensor responses across robots, which is difficult to guarantee with physical gas sensors.

Regarding coordination strategies, a wide range of approaches have been proposed for multi-robot information gathering, ranging from centralized \cite{nanavati_mrmste_2025,wiedemann_model-based_2019} to decentralized strategies~\cite{bourne_decentralized_2020} and learning-based methods~\cite{tzes_graph_2022}. In \cite{Ercolani_multi_robot_2024,rahbar_distributed_2020}, robots coordinate their navigation by independently selecting sampling locations, sharing these decisions, and exchanging goals to reduce the overall travel cost. 
 A graph neural network is proposed in \cite{tzes_graph_2022} that takes robot states and local estimations as input and outputs the next goal positions.
These methods rely on assigning discrete informative points to robots, typically following a stop-sense-go strategy \cite{jin_towards_2023}, where robots must stop at sampling locations to acquire measurements. While sense-in-motion approaches~\cite{jin_sense_2024} have been shown to lead to superior performance at the single robot level, in particular by enabling the robot to exploit measurements collected during motion, they introduce additional coordination challenges at the multi-robot level. In particular, partially overlapping trajectories can lead to redundant observations and reduced overall information gain. To reduce overlap and improve coverage of the \ac{MRS}, \cite{nanavati_distributed_2024} employs a Voronoi-based partitioning of the environment, assigning each robot to a distinct region for exploration.

Unlike existing works, where the homogeneous sensor response is assumed, and the path planning is generally designed for stop-sense-go strategy, in this work, we propose a distributed \ac{GSL} framework for \ac{MRS} equipped with heterogeneous gas sensors, while gathering samples during motion. Each robot independently estimates a local belief map based on its individual measurements using a rank-based gas feature that is invariant to sensor scaling and nonlinearities. The local beliefs are then fused through a \ac{PoE} formulation to obtain a global belief map. Based on the fused belief, we introduce an informative region allocation strategy that partitions the environment into spatially distinct regions, encouraging efficient exploration while reducing redundancy. Each robot then conducts the \ac{IPP} inside its assigned region. This approach enables the robot team to balance exploration and exploitation, and to effectively utilize measurements collected during motion. Building on our previously proposed single-robot rank-based \ac{STE} framework \cite{jin_uncalibrated_2026}, this paper presents the following contributions:
\begin{itemize}
    \item A distributed source estimation framework that enables robots equipped with uncalibrated sensors to independently estimate local posteriors and fuse them through a \ac{PoE} formulation.
    \item An informative region allocation and path planning strategy that promotes coordinated exploration and efficient information gathering.
    \item An extensive evaluation against a benchmark method in a high-fidelity simulation environment with realistic MOX sensor models under both calibrated and uncalibrated sensing conditions.  
\end{itemize}

\begin{figure}
\centering
\includegraphics[width=0.8\textwidth]{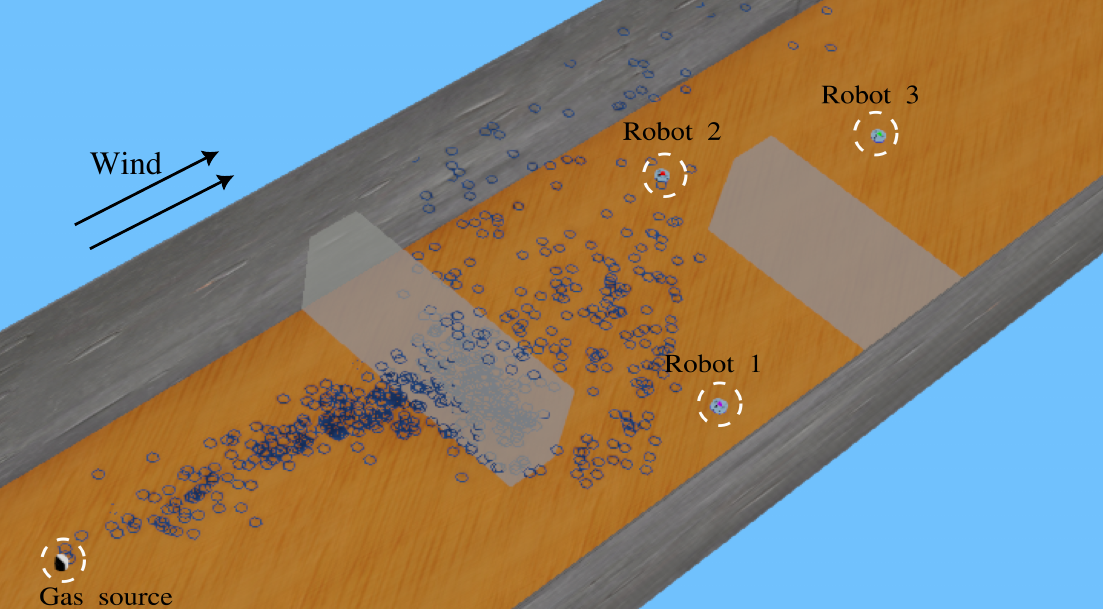}
\caption{A screenshot of the high-fidelity simulator Webots showing a \ac{MRS} engaged in a \ac{GSL} task.} \label{fig:simulation_setup}
\end{figure}

\section{Problem Statement}
Consider a two-dimensional indoor workspace $\Omega \subset \mathbb{R}^2$ containing obstacles $\mathcal{O}$, and assume that a single stationary gas source is located at an unknown position within the free space $\mathcal{F}=\Omega\setminus\mathcal{O}$. The workspace is discretized equally into $N_e$ grid cells. We define a discrete random variable $S\in \{1, ..., N_e\}$, where the event $S=i$ indicates that the gas source is located at the center of the $i$-th grid cell. A team of $M$ robots navigates in the environment and collects gas concentration measurements. Let $z_t^m$ representing the gas measurements gathered by robot $m\in\{1,...,M\}$ at time step $t$. Suppose the robot $m$ collected $N_k^m$ measurements up to iteration $k$. The corresponding measurement set is denoted as: $Z^m_k = \{z_1^m, z_2^m, ..., z_{N_k^m}^m\}$. 

The objective of \ac{GSL} is to estimate the posterior distribution of the source location based on collected measurements. At iteration $k$, the robot team maintains a belief map $b_k$ over the discretized environment as:
\begin{equation}
    b_k(i)=\Pr(S=i | Z^1_k, ..., Z^M_k), \qquad i =1,...,N_e,
\end{equation}
To quantify the uncertainty of the estimation, we use the entropy of the belief map:
    \begin{equation}
    H(b_k) = -\sum_{i=1}^{N_e} b_k(i)\,\log b_k(i).
\end{equation}
The estimated source location is given by the \ac{MAP} estimate:
\begin{equation}
        \hat{S}_k = \arg \max_{i \in \{1, \dots, N_e\}} b_k(i).
\end{equation}

The estimation process continues until either the entropy of the belief map falls below a predefined threshold or a maximum number of iterations is reached.

\section{Methodology}
\begin{figure}
    \centering
    \includegraphics[width=\linewidth]{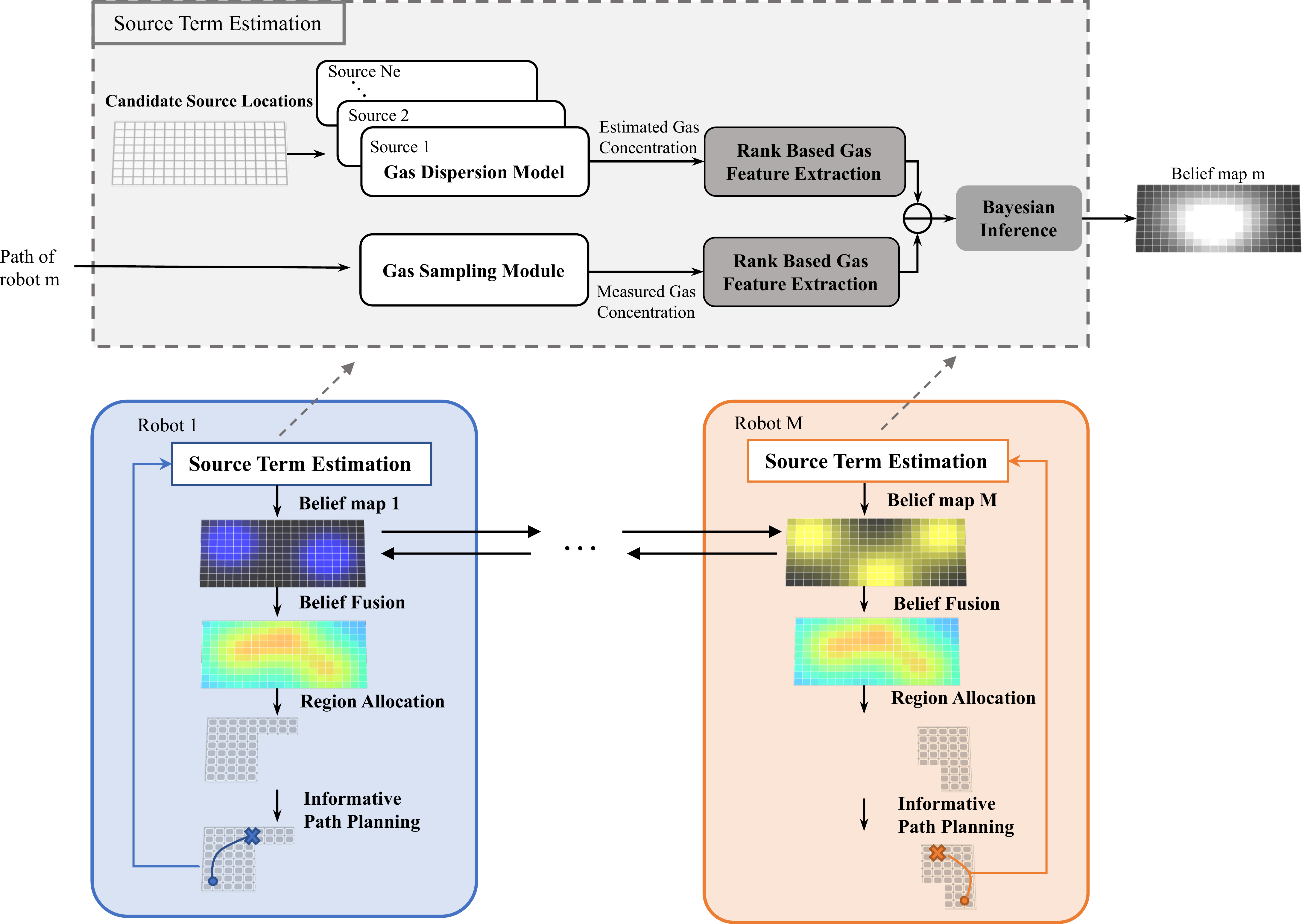}
    \caption{Workflow of the distributed \ac{STE} framework.}
    \label{fig:distributed_STE}
\end{figure}

We propose a distributed probabilistic framework for \ac{GSL} using a \ac{MRS} with heterogeneous sensors. The overall workflow is illustrated in Fig.~\ref{fig:distributed_STE}. Each robot independently estimates a belief map over the source location using its local observations within a \ac{STE} framework, which will be detailed in Sec.~\ref{sec:STE}. These local beliefs are then shared and fused among the group to form a global belief, which will be detailed in Sec.~\ref{sec:belief_sharing_merging}. The fused belief map is used to partition the map into $M$ non-overlapping regions, and then allocate one region to each robot. Each robot then plans its own informative path in its assigned region. The region allocation and \ac{IPP} will be detailed in Sec.~\ref{sec:region_allocation_IPP}.

At each iteration, robots collect measurements, update local beliefs, perform belief fusion, and replan their trajectories based on the updated global belief. The process terminates when the entropy of the global belief falls below a predefined threshold or the maximum iteration cycle is reached, and then the source location is declared.

\subsection{Source Term Estimation Framework}\label{sec:STE}
\ac{STE} is an inverse modeling approach used to infer parameters governing gas dispersion ~\cite{hutchinson_review_2017}. In this work, we formulate \ac{STE} as a Bayesian inference problem over the discrete source location variable $S$. This process incorporates a steady-state plume model and gas concentration measurements within the environment.
As shown in Fig.~\ref{fig:distributed_STE}, at each iteration, for every candidate source location $i$, a forward gas dispersion model predicts the expected gas concentration field $\hat{c}(\cdot|S=i)$ originating from this source. Given a robot's measurement set $Z_k^m$, both measured and predicted gas features are extracted and compared. The alignment between these features defines the likelihood of each source hypothesis, which is used to update the posterior belief over $S$. After reaching its goal location, each robot shares its local belief map with the team. Once all local beliefs are received, they are fused to obtain a global belief, which is then used to guide subsequent \ac{IPP}.

\subsubsection{Gas plume model - } To incorporate physical knowledge into the estimation process, we employ a gas plume model to predict concentrations under each source hypothesis. In complex indoor environments with obstacles, analytical models, such as the Pseudo-Gaussian plume model~\cite{holmes_review_2006}, are often too inaccurate. Therefore, we adopt a \ac{DDPM} from our previous work~\cite{jin_towards_2023} to predict the gas concentration field in built environments. 

This approach leverages a convolutional neural network that takes as input the environment configuration (e.g., inlet/outlet conditions and obstacle layout) and a candidate source location $S=i$, and outputs the predicted gas concentration field $\hat{c}(\cdot|S=i)$. Given a robot trajectory, predicted concentrations are sampled at robot positions: $\hat{c}_t^m(i) = \hat{c}(x_t^m|S=i)$, where $x_t^m$ is the position of robot $m$ at time step $t$.
For more information about \ac{DDPM} structure and
training process, please refer to \cite{jin_towards_2023}.
\subsubsection{Rank-based gas feature -} \label{sec: rank_based_gas_feature}
To estimate the likelihood of a hypothesis $S=i$, the alignment between the predicted gas concentration $\hat{c}_t^m(i)$ and the measured gas intensity $z_t^m$ is evaluated. However, direct comparison between predicted concentrations and sensor measurements is unreliable due to sensor heterogeneity and nonlinear responses, as well as the contrast between a  predicted value derived from a stationary plume model and a real-time measurement. To address this aspect, we use a rank-based feature proposed in our previous work \cite{jin_uncalibrated_2026} that captures the relative magnitude evolution of observations. 

For robot $m$, consider its measurement set $Z_k^m =\{z_1^m, ..., z_{N_k^m}^m\}$ up to iteration $k$. We compute the \ac{EDF} for the measurements:
\begin{equation}
\tilde{z}_t^m = \frac{1}{N_k^m} \sum_{j=1}^{N_k^m} \mathbf{1}_{z_j^m \leq z_t^m},
\end{equation}
which represents the normalized rank of each measurement $z_t^m$ in the set $Z_k^m$.

Similarly, for predicted concentrations:
\begin{equation}
\tilde{c}_t^m(i) = \frac{1}{N_k^m} \sum_{j=1}^{N_k^m} \mathbf{1}_{\hat{c}_j^m(i) \leq \hat{c}_t^m(i)}.    
\end{equation}
This representation enables comparison without requiring consistent scaling between measurements and predictions. It relies on the assumption that the sensor response is a monotonic transformation of the true gas concentration, which is generally satisfied for in-situ gas sensors despite their nonlinearities. Since ranking is computed independently for each robot, differences in sensor characteristics do not interfere across robots.

\subsubsection{Probabilistic distribution estimation -} Each robot evaluates the likelihood of its observations under each source hypothesis. We model the discrepancy between rank-based features using a Gaussian assumption, which provides a smooth and robust similarity measure while accounting for both measurement and model uncertainties. The likelihood for robot $m$ is defined as:
\begin{equation}
p^m(Z_{1:N_m^k}^m \mid S = i) \propto 
\exp \left(
- \frac{1}{2} \sum_{t=1}^{N_m^k} 
\frac{ \left( \tilde{c}_t^m(i) - \tilde{z}_t^m \right)^2 }
{\sigma_M^2 + \sigma_E^2}
\right),
\end{equation}
where $\sigma_M^2$ and $\sigma_E^2$ denote the measurement and model uncertainty, respectively.

Since the relative rank of each $z_t^m$ and $c_t^m$ will be adapted when new measurements are available, at each iteration $k$, all the historical measurements are used to update the local belief map: 
\begin{equation}
b_k^m(i) \propto p^m(Z_{1:N_m^k}^m \mid S = i) P(S=i),
\end{equation}
followed by normalization such that $\sum_{i=1}^{N_e} b_k^m(i)=1.$ The prior $p(S=i)$ is initialized as a uniform distribution over the free space of the environment, while cells occupied by obstacles are assigned zero probability.
\subsection{Belief Sharing and Merging} \label{sec:belief_sharing_merging}
After computing local belief maps $b_k^m(i)$, robots exchange and fuse them to obtain a global belief. Since all robots share the same forward model and environmental information, their beliefs are treated equally. We adopt a \ac{PoE} formulation~\cite{hinton_training_2002}:
\begin{equation}
    b_k(i) = \prod_{m=1}^Mb_k^m(i),
\end{equation}
followed by normalization, that $\sum_{i=1}^{N_e} b_k(i)=1$. While this formulation assumes conditional independence between robot observations, it remains effective in practice and emphasizes consensus across robots. A location will only achieve high probability if it is consistently supported by multiple robots, improving robustness to individual noise or bias.

\subsection{Informative Region Allocation and Path Planning} \label{sec:region_allocation_IPP}
Based on the fused belief $b_k$, robots plan trajectories to maximize collective information gain. Since measurements are collected continuously during motion, overlapping trajectories might bring limited additional information. Therefore, robots are encouraged to explore spatially distinct regions. At the same time, the rank-based feature requires each robot to sample a sufficiently diverse range of gas concentrations, creating a trade-off between global exploration and local exploitation. To address this, we design a coordination strategy that gradually guides the robot team from exploration to exploitation.
\begin{figure}
    \centering
    \includegraphics[width=\linewidth]{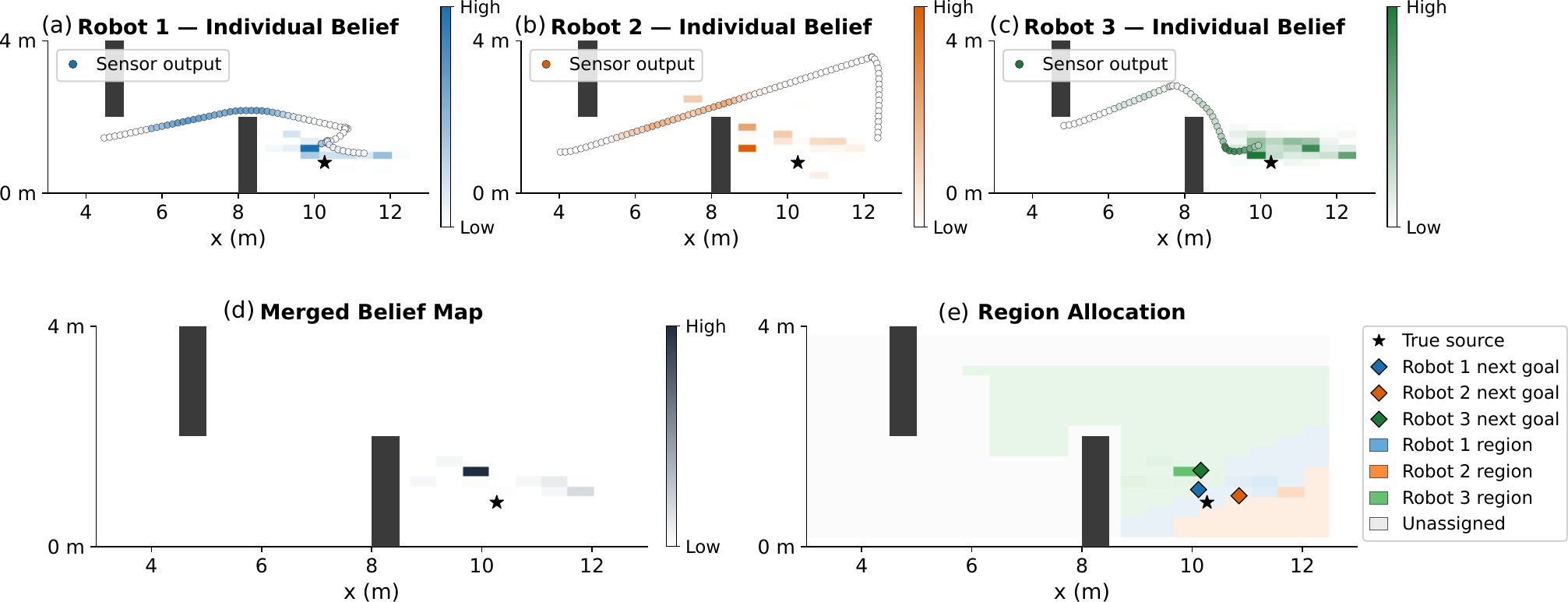}
    \caption{One iteration of distributed estimation and coordination: (a)-(c) robot trajectories, sensor outputs, and corresponding local beliefs; (d) global belief after PoE fusion; and (e) non-overlapping region allocation and information-weighted navigation goals. The star denotes the true source location.}
    \label{fig:belief_clusters}
\end{figure}
\subsubsection{Region allocation method - }The goal of region allocation is to spatially distribute robots to maximize coverage while prioritizing high-probability areas. To this end, the environment is partitioned into $M$ spatially connected, non-overlapping regions, with each region assigned to one robot.

We first identify $N_{\text{active}}$  active cells whose belief value satisfies $b_i>\epsilon$, where $\epsilon$ is a threshold indicating sufficiently informative locations. From these, the top-$M$ highest-probability cells are selected as seeds to initialize the regions. Each seed is assigned to the nearest robot. Regions are then expanded using a multi-seed breadth-first search on an 8-connected grid. To maintain balanced region sizes, expansion prioritizes smaller regions and enforces a soft size constraint:
\begin{equation}
    |C_m| \leq \alpha \cdot \frac{N_{\text{active}}}{M}
\end{equation}
where $C_m$ is the set of cells assigned to robot $m$ and $\alpha > 1$ is a relaxation factor. A cell is assigned to a region only if it is not an obstacle and has not been previously assigned. This ensures spatial connectivity, balanced workload, and no overlap between regions.

\subsubsection{Path planning within each region - }
Each robot plans its trajectory within its assigned region $C_m$. Instead of targeting a single cell, the robot moves toward an information-weighted centroid:
\begin{equation}
x_{\mathrm{info}}^m = \frac{\sum_{x_i \in C_m} b_k(i)\, x_i}{\sum_{x_i \in C_m} b_k(i)}.    
\end{equation}
This centroid-based strategy provides a computationally efficient approximation of information-driven exploration, balancing coverage and exploitation. In early stages, when the belief distribution is diffuse, it promotes broad exploration. As the belief becomes more concentrated, the centroid shifts toward high-probability regions, encouraging focused exploitation. 

Fig.~\ref{fig:belief_clusters} illustrates one planning iteration. Panels (a)-(c) show the trajectory and sensor output of each robot together with the local belief map estimated from its own measurement history. Darker trajectory markers denote higher sensor outputs, while darker belief map cells denote higher source probabilities. Panel (d) shows the global belief obtained by PoE fusion. The fusion suppresses hypotheses that are not consistently supported across the local beliefs, thereby concentrating the belief on a smaller candidate region. Panel (e) shows the resulting allocation of the active cells to the robots. The diamond in each region denotes its weighted information centroid, which is used as the robot's next navigation goal.

\section{Experiments and Results}
In this section, we introduce the experimental setup and the simulated heterogeneous \ac{MOX} sensors and benchmark the proposed method with a baseline STE algorithm.
\subsection{Simulation Setup}
We evaluate the proposed method in a high-fidelity simulation environment using Webots~\cite{Webots}, integrated with a dedicated gas dispersion plugin \cite{webots_odor}. A team of three Khepera IV robots, each equipped with a \ac{MOX} gas sensor, navigate the environment and collect gas concentration measurements. An example of the simulation setup is shown in Fig.~\ref{fig:simulation_setup}. 

\begin{figure}
    \centering
    \includegraphics[width=\linewidth]{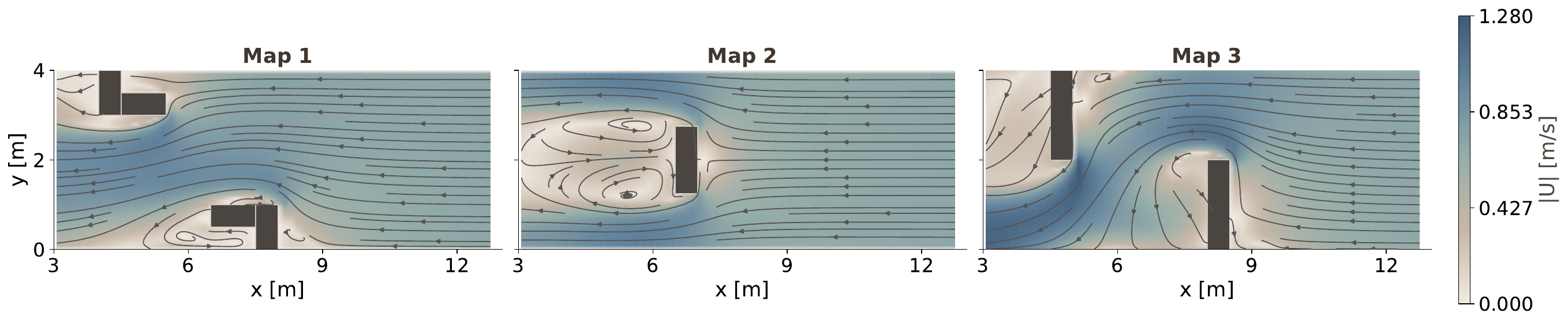}
    \caption{Experimental environments used for evaluation. Three indoor maps with different obstacle configurations are shown, along with the corresponding wind fields.}
    \label{fig:test_maps}
\end{figure}

The experiments are conducted in three indoor environments with different obstacle configurations, as illustrated in Fig.~\ref{fig:test_maps}. These obstacles create complex airflow patterns that significantly affect gas dispersion. In all scenarios, a steady airflow is applied from the right (inlet) to the left (outlet). The gas source is placed near the inlet at a random location, while robots are initialized near the outlet with random starting positions.

\subsubsection{Heterogeneous MOX sensor simulation - } \label{sec:heterogenous_sensor}
To realistically model sensor heterogeneity, we adopt the MOX sensor model from our previous work \cite{jin_uncalibrated_2026}, incorporating the three following key aspects. 

\begin{figure} \centering \includegraphics[width=0.8\linewidth]{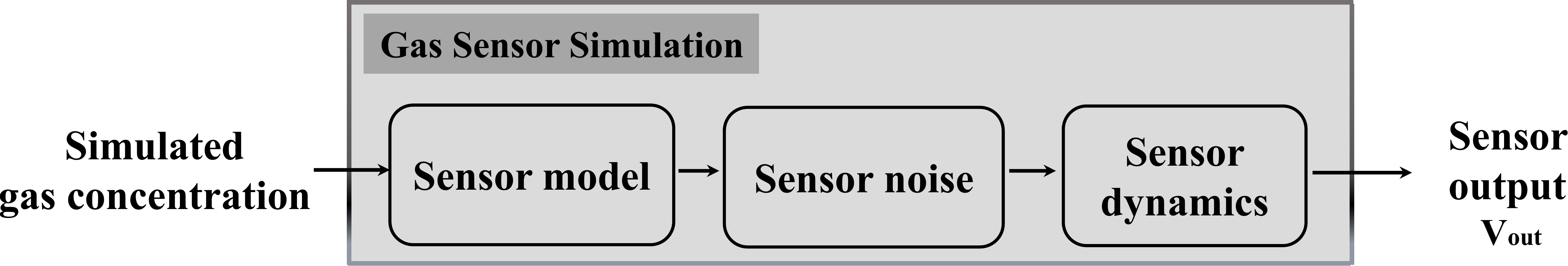} \caption{Data flow of the MOX sensor simulation. The ground-truth gas concentration is transformed through a nonlinear sensor model, followed by additive noise and dynamic response modeling, to generate realistic sensor outputs.} \label{fig:MOX_sensor_simulation} \end{figure}

\begin{itemize}
    \item Nonlinear response: the sensor output voltage is a nonlinear function of gas concentration, parameterized by a baseline resistance, which varies across sensors due to manufacturing differences, and is sensitive to the environmental conditions.
    \item Sensor noise: additive noise consists of a constant background component and a signal-dependent intermittency component, reflecting real gas sensing conditions.
    \item Slow dynamics: response and recovery behavior are modeled as a first-order dynamic system, with parameters calibrated with a real MOX sensor.
\end{itemize}
To simulate heterogeneous sensing, a different baseline resistance (100 $\mathrm{\Omega}$, 500 $\mathrm{\Omega}$ and 1500 $ \mathrm{\Omega}$) is assigned to each robot, covering the typical variability range observed in real sensors. The overall data flow of the sensor simulation is illustrated in Fig.~\ref{fig:MOX_sensor_simulation}. For more details about sensor modeling, please refer to~\cite{jin_uncalibrated_2026}.

In our setting, the gas sensor samples at 1 Hz and the maximal robot speed is set to 0.27 m/s. Real gas plumes are inherently discontinuous and intermittent, resulting in rapidly fluctuating gas concentrations. During continuous sampling, the slow sensor dynamics effectively act as a low-pass filter, smoothing rapid concentration fluctuations. Therefore, we directly use the output of the sensor to extract rank-based features without further signal processing.

\subsection{Performance Evaluation}
\subsubsection{Benchmark method - }We compare the proposed approach, referred to as Belief Sharing in the following, against a baseline algorithm, which we call Measurement Aggregation, that leverages the same STE framework but aggregates all measurements of the robots into a single dataset. In this baseline algorithm, each robot aggregate the gas measurements gathered by all the robots, and use the entire gas measurement dataset to compare with the model prediction to update the source belief. Unlike our method, the baseline algorithm assumes that sensor measurements are directly comparable across robots and does not account for the actual sensor heterogeneity and nonlinear response, which is a common assumption adopted in existing work. For the sake of benchmarking fairness, the same region allocation and path planning strategies are also adopted for the baseline algorithm.

To benchmark the performance of Belief Sharing with that of Measurement Aggregation, we evaluate the algorithms under two different settings:
\begin{itemize}
    \item Calibrated sensors: the gas sensors of the robots are all calibrated towards the gas plume model and can directly measure the ground truth gas concentration.
    \item Uncalibrated sensors: the gas sensors are not calibrated and are simulated with the model introduced in Sec.~\ref{sec:heterogenous_sensor}; the raw sensor output is directly used for the experiments.
\end{itemize}

For each setting, we evaluate both algorithms under three maps. For each environment, 30 trials with randomized source and robot initial positions are conducted, resulting in a total of 360 simulation runs.

\subsubsection{Evaluation metrics - }
The performance of the \ac{MRS} is evaluated in terms of both localization accuracy and efficiency as follows.
\begin{itemize}
    \item Localization accuracy: measured as the Euclidean distance between the estimated source location (MAP estimate) and the ground-truth source position.
    \item Exploration efficiency: quantified by the total trajectory length traveled by the robot team until end of the task.
\end{itemize}

\subsubsection{Results}
\begin{figure}
    \centering
    \includegraphics[width=\linewidth]{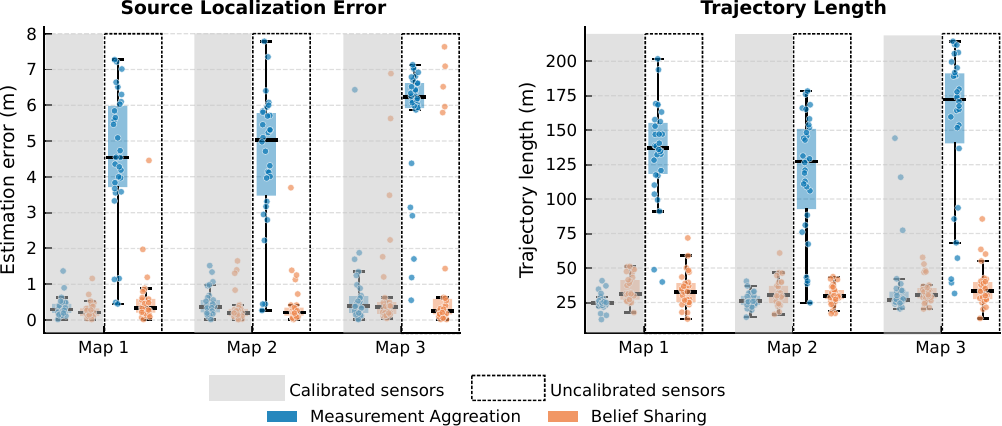}
    \caption{Performance comparison between the Benchmark (measurement aggregation) and the proposed method (belief sharing) across three environments.}
    \label{fig:simulation_results}
\end{figure}

The performance comparison is shown in Fig.~\ref{fig:simulation_results}. The gray zones show the results with calibrated sensors and the zones with dashed bounding boxes are the results with uncalibrated sensors.

With calibrated sensors, Belief Sharing and Measurement Aggregation achieve comparable localization accuracy and trajectory lengths, since measurements are consistent across robots. However, in the uncalibrated sensors settings, the baseline algorithm consistently fails to localize the source across all map configurations, yielding large estimation errors despite addition traveling (and therefore sampling). This is primarily due to the mismatch between predicted concentrations and raw sensor measurements, caused by sensor heterogeneity and nonlinear responses. Without proper calibration, aggregating measurements across robots leads to inconsistent likelihood estimation and prevents convergence.

In contrast, the Belief Sharing algorithm successfully localizes the gas source in most cases across all environments regardless of calibrated or uncalibrated sensors. By using rank-based features, each robot estimates its local belief based on relative ordering of measurements rather than their absolute values, effectively removing the need for sensor calibration. The subsequent belief fusion reinforces consistent hypotheses across robots. In terms of efficiency, the proposed method achieves source localization with moderate trajectory lengths, demonstrating a good balance between exploration and exploitation.

Overall, the results demonstrate that the proposed distributed estimation framework is robust to heterogeneous sensor responses and enables reliable \ac{GSL} in complex environments.

\subsection{Scalability of the Proposed Method}
In the current all-to-all communication architecture, each robot shares an $N_e$-dimensional belief map with the team at every planning iteration, resulting in a total communication cost of $\mathcal{O}(M^2N_e)$. Belief fusion at each robot requires $\mathcal{O}(MN_e)$ operations. The region allocation process scales approximately with the number of active cells, while the \ac{STE} and \ac{IPP} are performed locally and do not scale with the team size. Overall, the framework is suitable for small- to medium-sized robot teams, while more communication-efficient belief fusion strategies could further improve scalability.

\section{Conclusion and Outlook}
In this work, we presented a distributed estimation framework for \ac{GSL} using a \ac{MRS} equipped with uncalibrated gas sensors. The proposed approach enables each robot to independently estimate a local belief using rank-based gas features, which are invariant to sensor scaling and nonlinearities. These local beliefs are then fused to obtain a global estimate of the source location.

To support efficient coordination, we introduce an informative region allocation and path planning strategy that balances exploration and exploitation while reducing redundant observations. The framework allows robots to continuously collect measurements during motion and collaboratively refine their belief over the source location.

Extensive simulations in a high-fidelity environment with realistic \ac{MOX} sensor models demonstrate that the proposed method significantly outperforms a baseline algorithm that aggregates raw sensor measurements. In particular, our approach achieves reliable localization despite strong sensor heterogeneity, highlighting the importance of calibration-free features in distributed estimation. 

Although this work focuses on \ac{GSL} as a case study, the proposed framework is more general and can be applied to a broader class of distributed estimation problems in \ac{MRS}. Future work will focus on improving communication efficiency by reducing the amount of data exchanged between robots, for example by transmitting only incremental updates of the belief map. Additionally, extending the framework to real-world experiments remains an important direction for further validation.

%
%

\begin{credits}
\subsubsection{\ackname} This work is funded by the Swiss National Science
Foundation under grants 200020\_175809 and 10.001.747. 
\subsubsection{Further information.} Additional information about the
research can be found here: \url{https://disal.epfl.ch/research/gassensingstructure/}.
\subsubsection{\discintname} The authors declare no conflict of interest.
\end{credits}
%
%
%
\bibliographystyle{splncs04}
\bibliography{IEEEexample}

@misc{webots_odor,
	title = {Webots {Odor} {Simulation}/{Model} - {Wikibooks}, open books for an open world},
	url = {https://en.wikibooks.org/wiki/Webots\_Odor\_Simulation/Model},
	language = {en},
	urldate = {2022-03-22},
}

@article{hutchinson_review_2017,
	title = {A review of source term estimation methods for atmospheric dispersion events using static or mobile sensors},
	volume = {36},
	issn = {15662535},
	language = {en},
	urldate = {2021-10-05},
	journal = {Information Fusion},
	author = {Hutchinson, Michael and Oh, Hyondong and Chen, Wen-Hua},
	month = jul,
	year = {2017},
	pages = {130--148},
}

@inproceedings{rahbar_distributed_2020,
	title = {A {Distributed} {Source} {Term} {Estimation} {Algorithm} for {Multi}-{Robot} {Systems}},
	isbn = {978-1-72817-395-5},
	language = {en},
	urldate = {2021-08-31},
	booktitle = {{IEEE} {International} {Conference} on {Robotics} and {Automation}},
	author = {Rahbar, Faezeh and Martinoli, Alcherio},
	month = may,
	year = {2020},
	pages = {5604--5610},
}

@article{Webots,
  title={Cyberbotics {{L}}td. {{W}}ebots™: professional mobile robot simulation},
  author={Michel, Olivier},
  journal={International Journal of Advanced Robotic Systems},
  volume={1},
  number={1},
  pages={39--42},
  year={2004},
  publisher={SAGE Publications Sage UK: London, England}
}

@article{bourne_decentralized_2020,
	title = {Decentralized {Multi}-agent information-theoretic control for target estimation and localization: finding gas leaks},
	volume = {39},
	issn = {0278-3649, 1741-3176},
	shorttitle = {Decentralized {Multi}-agent information-theoretic control for target estimation and localization},
	language = {en},
	number = {13},
	urldate = {2022-07-29},
	journal = {The International Journal of Robotics Research},
	author = {Bourne, Joseph R and Goodell, Matthew N and He, Xiang and Steiner, Jake A and Leang, Kam K},
	month = nov,
	year = {2020},
	pages = {1525--1548},
}

@inproceedings{jin_towards_2023,
	title = {Towards {Efficient} {Gas} {Leak} {Detection} in {Built} {Environments}: {Data}-{Driven} {Plume} {Modeling} for {Gas} {Sensing} {Robots}},
	isbn = {9798350323658},
	shorttitle = {Towards {Efficient} {Gas} {Leak} {Detection} in {Built} {Environments}},
	language = {en},
	urldate = {2023-07-27},
	booktitle = {{IEEE} {International} {Conference} on {Robotics} and {Automation}},
	author = {Jin, Wanting and Rahbar, Faezeh and Ercolani, Chiara and Martinoli, Alcherio},
	month = may,
	year = {2023},
	pages = {7749--7755},
}

@article{wiedemann_model-based_2019,
	title = {Model-based gas source localization strategy for a cooperative multi-robot system—{A} probabilistic approach and experimental validation incorporating physical knowledge and model uncertainties},
	volume = {118},
	issn = {09218890},
	language = {en},
	urldate = {2021-06-15},
	journal = {Robotics and Autonomous Systems},
	author = {Wiedemann, Thomas and Shutin, Dmitriy and Lilienthal, Achim J.},
	month = aug,
	year = {2019},
	pages = {66--79},
}

@article{holmes_review_2006,
	title = {A review of dispersion modelling and its application to the dispersion of particles: {An} overview of different dispersion models available},
	volume = {40},
	issn = {13522310},
	shorttitle = {A review of dispersion modelling and its application to the dispersion of particles},
	language = {en},
	number = {30},
	urldate = {2022-08-31},
	journal = {Atmospheric Environment},
	author = {Holmes, N.S. and Morawska, L.},
	month = sep,
	year = {2006},
	pages = {5902--5928},
}

@article{nanavati_distributed_2024,
	title = {Distributed multi-robot source term estimation with coverage control and information theoretic based coordination},
	volume = {111},
	issn = {15662535},
	language = {en},
	urldate = {2024-06-18},
	journal = {Information Fusion},
	author = {Nanavati, Rohit V. and Coombes, Matthew J. and Liu, Cunjia},
	month = nov,
	year = {2024},
	pages = {102503},
}

@inproceedings{jin_sense_2024,
	title = {Sense in {Motion} with {Belief} {Clustering}: {Efficient} {Gas} {Source} {Localization} with {Mobile} {Robots}},
	copyright = {https://doi.org/10.15223/policy-029},
	isbn = {9798350384574},
	shorttitle = {Sense in {Motion} with {Belief} {Clustering}},
	language = {en},
	urldate = {2024-12-05},
	booktitle = {{IEEE} {International} {Conference} on {Robotics} and {Automation}},
	author = {Jin, Wanting and Martinoli, Alcherio},
	month = may,
	year = {2024},
	pages = {14909--14916},
}

@article{hinton_training_2002,
	title = {Training {Products} of {Experts} by {Minimizing} {Contrastive} {Divergence}},
	volume = {14},
	issn = {0899-7667, 1530-888X},
	language = {en},
	number = {8},
	urldate = {2026-04-28},
	journal = {Neural Computation},
	author = {Hinton, Geoffrey E.},
	month = aug,
	year = {2002},
	pages = {1771--1800},

}

@INPROCEEDINGS{Ercolani_multi_robot_2024,
  author={Ercolani, Chiara and Deshmukh, Shashank Mahendra and Peeters, Thomas Laurent and Martinoli, Alcherio},
  booktitle={IEEE International Conference on Robotics and Automation}, 
  title={Multi-Robot 3D Gas Distribution Mapping: Coordination, Information Sharing and Environmental Knowledge}, 
  year={2023},
  volume={},
  number={},
  pages={11418-11424},
}

@article{park_cooperative_2020,
	title = {Cooperative information-driven source search and estimation for multiple agents},
	volume = {54},
	issn = {15662535},
	language = {en},
	urldate = {2026-04-29},
	journal = {Information Fusion},
	author = {Park, Minkyu and Oh, Hyondong},
	month = feb,
	year = {2020},
	pages = {72--84},
}

@misc{tzes_graph_2022,
	title = {Graph {Neural} {Networks} for {Multi}-{Robot} {Active} {Information} {Acquisition}},
	language = {en},
	urldate = {2023-07-11},
	publisher = {arXiv},
	author = {Tzes, Mariliza and Bousias, Nikolaos and Chatzipantazis, Evangelos and Pappas, George J.},
	month = sep,
	year = {2022},
}

@misc{nanavati_mrmste_2025,
	title = {Mr.{MSTE}: {Multi}-robot {Multi}-{Source} {Term} {Estimation} with {Wind}-{Aware} {Coverage} {Control}},
	shorttitle = {Mr.{MSTE}},
	language = {en},
	urldate = {2026-02-25},
	publisher = {arXiv},
	author = {Nanavati, Rohit V. and Glover, Tim J. and Coombes, Matthew J. and Liu, Cunjia},
	month = dec,
	year = {2025},
}

@inproceedings{jin_uncalibrated_2026,
	title = {Calibration-Free Gas Source Localization with Mobile Robots: Source Term Estimation Based on Concentration Measurement Ranking},
	language = {en},
	booktitle = {{IEEE} {International} {Conference} on {Robotics} and {Automation}},
    author = {Jin, Wanting and Duranceau, Agatha and
          Er{\"u}nsal, {\.I}zzet Ka{\u{g}}an and
          Martinoli, Alcherio},
	month = jun,
	year = {2026},
}

@incollection{park2023information,
  title={Information-Theoretic Autonomous Source Search and Estimation of Mobile Sensors},
  author={Park, Minkyu and An, Seulbi and Jang, Hongro and Oh, Hyondong},
  booktitle={Control of Autonomous Aerial Vehicles: Advances in Autopilot Design for Civilian UAVs},
  pages={135--166},
  year={2023},
  publisher={Springer}
}

@article{tian2022kimera,
  title={Kimera-multi: Robust, distributed, dense metric-semantic slam for multi-robot systems},
  author={Tian, Yulun and Chang, Yun and Arias, Fernando Herrera and Nieto-Granda, Carlos and How, Jonathan P and Carlone, Luca},
  journal={IEEE Transactions on Robotics},
  volume={38},
  number={4},
  year={2022},
  publisher={IEEE}
}

@incollection{jung2006cooperative,
  title={Cooperative multi-robot target tracking},
  author={Jung, Boyoon and Sukhatme, Gaurav S},
  booktitle={Distributed Autonomous Robotic Systems 7},
  pages={81--90},
  year={2006},
  publisher={Springer}
}

@inproceedings{werner2024autonomous,
  title={Autonomous localization of multiple ionizing radiation sources using miniature single-layer Compton cameras onboard a group of micro aerial vehicles},
  author={Werner, Michal and B{\'a}{\v{c}}a, Tom{\'a}{\v{s}} and {\v{S}}tibinger, Petr and Doubravov{\'a}, Daniela and {\v{S}}olc, Jaroslav and Rus{\v{n}}{\'a}k, Jan and Saska, Martin},
  booktitle={IEEE/RSJ International Conference on Intelligent Robots and Systems},
  pages={5710--5717},
  month = Oct,
  year={2024},
}

@article{rascon2017localization,
  title={Localization of sound sources in robotics: A review},
  author={Rascon, Caleb and Meza, Ivan},
  journal={Robotics and Autonomous Systems},
  volume={96},
  pages={184--210},
  year={2017},
  publisher={Elsevier}
}

@ARTICLE{Miao_Wang_2024,
  author={Wang, Miao and Xin, Bin and Jing, Mengjie and Qu, Yun},
  journal={IEEE Transactions on Automation Science and Engineering}, 
  title={A Priority-Based Multi-Robot Search Algorithm for Indoor Source Searching}, 
  year={2025},
  volume={22},
  number={},
  pages={10457-10469},
}

\end{document}